\pdfoutput=1

\documentclass[11pt]{article}

\usepackage[final]{coling}

\usepackage{times}
\usepackage{latexsym}

\usepackage[T1]{fontenc}

\usepackage[utf8]{inputenc}

\usepackage{microtype}

\usepackage{inconsolata}

\usepackage{graphicx}
\usepackage{listings}
\usepackage{tcolorbox}
\usepackage{amssymb}
\usepackage{array}
\definecolor{myLightGray}{rgb}{0.95, 0.95, 0.95}

\title{SyntheT2C: Generating Synthetic Data for Fine-Tuning Large Language Models on the Text2Cypher Task}

\author{Zijie Zhong \\
  Shanghai AI Laboratory \\
  \texttt{zhongzijie@pjlab.org.cn} \\\And
  Linqing Zhong \\
  Beihang University \\
  \texttt{lqzhong@buaa.edu.cn} \\\And
  Zhaoze Sun \\
  Beihang University \\
  \texttt{szz20241050@buaa.edu.cn} \\\AND
  Qingyun Jin \\
  Beihang University \\
  \texttt{sy2303108@buaa.edu.cn} \\\And
  Zengchang Qin\footnotemark[1] \\
  Beihang University \\
  \texttt{zcqin@buaa.edu.cn} \\\And
  Xiaofan Zhang\footnotemark[1] \\
  Shanghai AI Laboratory \\
  \texttt{zhangxiaofan@pjlab.org.cn} \\}

\begin{document}
\tcolorboxenvironment{lstlisting}{colback=myLightGray,colframe=black}

\maketitle
\footnotetext[1]{Corresponding authors}
\begin{abstract}
 Integrating Large Language Models (LLMs) with existing Knowledge Graph (KG) databases presents a promising avenue for enhancing LLMs' efficacy and mitigating their ``hallucinations''. Given that most KGs reside in graph databases accessible solely through specialized query languages (e.g., Cypher), it is critical to connect LLMs with KG databases by automating the translation of natural language into Cypher queries (termed as ``Text2Cypher'' task). Prior efforts tried to bolster LLMs' proficiency in Cypher generation through Supervised Fine-Tuning (SFT). However, these explorations are hindered by the lack of annotated datasets of Query-Cypher pairs, resulting from the labor-intensive and domain-specific nature of such annotation. In this study, we propose \textbf{SyntheT2C}, a methodology for constructing a synthetic Query-Cypher pair dataset, comprising two distinct pipelines: (1) LLM-based prompting and (2) template-filling. SyntheT2C is applied to two medical KG databases, culminating in the creation of a synthetic dataset, \textbf{MedT2C}. Comprehensive experiments demonstrate that the MedT2C dataset effectively enhances the performance of backbone LLMs on Text2Cypher task via SFT. Both the SyntheT2C codebase and the MedT2C dataset are released in \href{https://github.com/ZGChung/SyntheT2C}{https://github.com/ZGChung/SyntheT2C}.
\end{abstract}

\section{Introduction}

\begin{figure}[t]
  \includegraphics[width=\columnwidth]{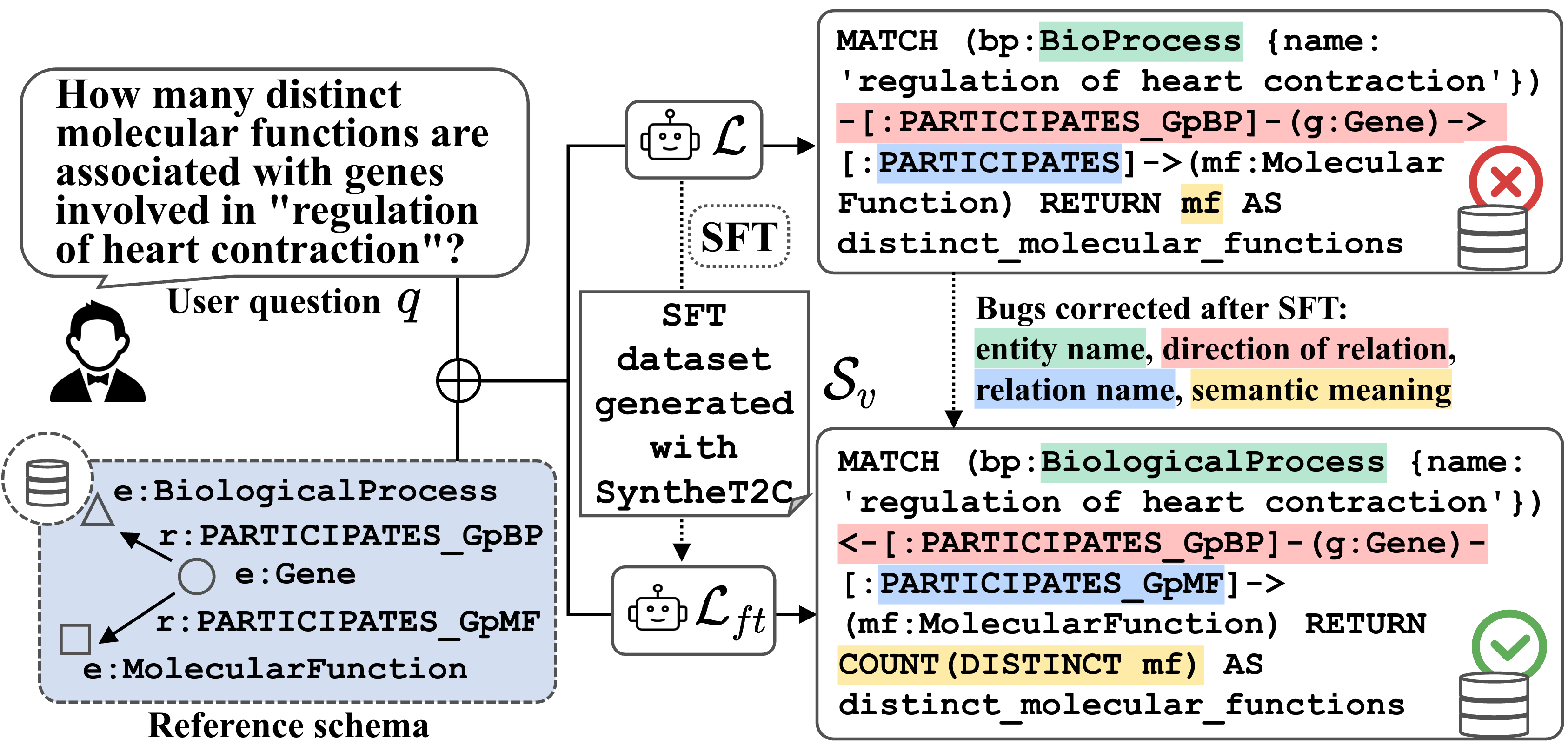}
  \caption{SyntheT2C builds synthetic data with two pipelines to SFT LLMs so that their performance on Text2Cypher task is enhanced.}
  \label{fig:showcase}
\end{figure}

Knowledge Graphs (KGs) constitute vital reservoirs of information within the Retrieval-Augmented Generation (RAG) paradigm \citep{rag} of Large Language Models (LLMs). Distinguished from other information sources, KGs boast structured and meticulously curated data, rendering them conducive to seamless updates and rectifications. Such attributes position KGs as pivotal instruments for mitigating issues of knowledge cutoff and ``hallucinations'' within LLMs. The inherent fidelity and adaptability of KGs make them practical assets for numerous knowledge-intensive products and applications \citep{finkg, hkgreview, pkg}. With the advent of LLMs, many researchers have focused on synergizing KGs with LLMs following the RAG framework, catapulting KGs to the forefront of academic research.

The efficient utilization of KG remains a formidable challenge because of the difference in format. Early methodologies involve direct extraction of triplets from KGs, subsequently integrating these text-form triplets directly into the prompts of LLMs \citep{talklikeagraph}. However, this approach often fails to concurrently preserve both semantic and structural nuances inherent within the KG. An alternative approach involves querying existing graph databases just like human users, promising accurate and interpretable results. Nonetheless, the primary impediment lies in the LLM's ability to formulate correct and precise queries. To address this limitation, numerous query generation tools or methodologies \citep{crake, dtqa, spsql} are proposed, aiming to translate human users' requests (in natural language) into query languages. This task assumes paramount importance for two pivotal reasons: (1) it empowers LLMs to consistently produce reliable queries, thereby enabling them to address knowledge deficits via direct interaction with KG databases; (2) it significantly facilitates human users' interaction with KG databases because learning the specific query language is no longer necessary. Among the spectrum of query generation research, the sub-task of translating natural language into the Cypher \citep{cypher} query language for Neo4j \citep{neo4j} databases stands out as a prominent research focus for two factors: (1) Neo4j is a widely adopted solution for KG databases, positioning Cypher as an essential tool for accessing these extensive repositories; (2) Cypher is a query language specifically designed for querying graph structures, offering significantly faster performance than other query languages, such as SQL, when processing graph data. Consequently, our work centers on this sub-task, commonly termed as ``\textbf{T}ext\textbf{2C}ypher'' (T2C).

A similar task to the Text2Cypher task is the ``Text2SQL'' task, wherein researchers endeavor to translate natural language sentences into SQL queries. Leveraging manually annotated datasets like SPIDER \citep{spider}, numerous methodologies have emerged, including SpCQL \citep{spcql} and SQLNet \citep{sqlnet}. Conversely, scant attention has been directed towards the Text2Cypher task. Existing approaches typically resort to decomposing a complete query into smaller components and translating each part separately. For instance, $R^3$-NL2GQL \citep{r3nl2gql} partitions the query generation process into CRUD keywords prediction, clause selection, and object type identification. Despite the success of these methods, adapting them to a specific KG database demands substantial extra effort. With the rise of LLMs, using LLMs for Cypher query generation appears promising. Notably, to the best of our knowledge, no endeavors have explored the potential application of LLMs to the Text2Cypher task. Our work aims to bridge this gap in the literature.

The Cypher writing performance of vanilla LLMs is not satisfactory. To improve it, we employ SFT, which necessitates a dataset of Question-Cypher pairs. However, creating such a dataset is challenging as it requires both domain-specific knowledge of the KG's content and expertise in Cypher's syntax. Consequently, there is currently no annotated dataset for the Text2Cypher task. To overcome this obstacle, we introduce SyntheT2C, a method designed to produce high-quality synthetic Question-Cypher pairs through two distinct pipelines: LLM-based prompting and template-filling (as shown in Figure~\ref{fig:showcase}). The LLM-based prompting pipeline aims to generate Cypher queries with greater semantic flexibility, while the template-filling pipeline focuses on producing syntactically complex Cypher queries. The generated Question-Cypher pairs undergo rigorous automated and manual validation, before being used to fine-tune backbone LLMs. The performance of Cypher generation is evaluated with a manually annotated evaluation dataset, complemented by a qualitative assessment using GPT as a judge. Additionally, we conduct a scalability test by fine-tuning the LLMs with larger synthetic datasets, which demonstrates that the synthetic data generated using our method does not collapse into simple patterns, thereby establishing the robustness of our approach for larger-scale applications.

SyntheT2C is tested with two medical KG databases: the LHY database and the Hetionet database (details in Section~\ref{sec:databases}). The generated synthetic dataset, ``MedT2C'', will be made public.

In conclusion, our main contributions are:

(1) We propose the SyntheT2C framework containing two pipelines to build synthetic datasets with any Neo4j database. Our method can generate Cypher that are both grammatically correct and syntactically diverse, facilitating the construction of SFT datasets.

(2) We test and validate the effectiveness and scalability of the synthetic dataset generated with SyntheT2C. The LLMs after fine-tuning show improved Cypher writing abilities. 

(3) We opensource a synthetic dataset MedT2C of optimal size, ready to be used for SFT.

\section{Related works}
\subsection{Knowledge Graph and graph database}
KGs have emerged as fundamental resources for organizing, representing, and querying vast amounts of interconnected information or domain-specific knowledge. These graphs find applications across various domains, including but not limited to, healthcare \citep{hkgreview, hkgconstruction}, finance \citep{finkgpipeline, finkg}, and e-commerce \citep{pkg}. In the realm of Natural Language Processing, KGs serve as invaluable sources of context and factual knowledge, enabling systems to reason, infer, and generate responses with enhanced accuracy and coherence. To handle the processing of graph data, a series of graph databases were invented, including Neo4j \citep{neo4j}, NebulaGraph \citep{nebula}, and Amazon Neptune \citep{neptune}. Among them, our work focuses on the Neo4j database \citep{neo4j}, a widely adopted graph database management system. Neo4j database employs Cypher query language for expressing complex graph patterns during the retrieval. 

\subsection{Large Language Models}
LLMs are advanced AI models that have been trained on vast amounts of text data to understand and generate human-like language. Following the milestone release of InstructGPT \citep{instructgpt} by OpenAI, a series of LLMs are built, featuring different advantages and drawbacks, e.g., the series of GPT models \citep{gpt3, gpt4} by OpenAI, Llama \citep{llama3} by Meta, Qwen \citep{qwen} by Alibaba Cloud, InternLM \citep{internlm2} by Shanghai AI Lab, etc. Recent researches highlight LLMs' ability to utilize external existing tools like calculator, search engine, or databases \citep{gorilla, webgpt, llmastoolmaker, toolllm}, which is usually abstracted as ``Function calling''. Many of its implementations involve generating codes or queries to interact with external tools. 

\subsection{Code generation}
Code Generation is the process of automatically producing executable code from a higher-level representation or natural language. With the advent of LLMs, code generation has experienced a significant advancement. LLMs can now be trained on vast amounts of code and programming-related text materials, enabling them to understand and generate code snippets based on given requirements (e.g., Codex \citep{codex}, Polycoder \citep{polycoder}, and Code Llama \citep{codellama}). By leveraging the contextual understanding \citep{icl} and language capabilities of LLMs, code generation becomes more efficient, accurate, and adaptable. Code generation with LLM is not only useful in helping developers to write codes but also in providing a powerful ``language'' for LLM to interact with other tools: LLMs can be tuned to output executable codes or queries to manipulate external resources. This is the fundamental idea for research in ``Function Calling'' and Multi-Agent Systems. Current code generation methods rely on two methods for evaluation: either with automatic metrics calculated with an annotated evaluation dataset \citep{bleu, rouge, meteor, codebleu, codebertscore} or with comparison by a judge (human or powerful LLM like GPT-4) \citep{llmasajudge}. Both evaluation methods are used in our work.

\section{Methodology}

\begin{figure*}[t]
  \includegraphics[width=\linewidth]{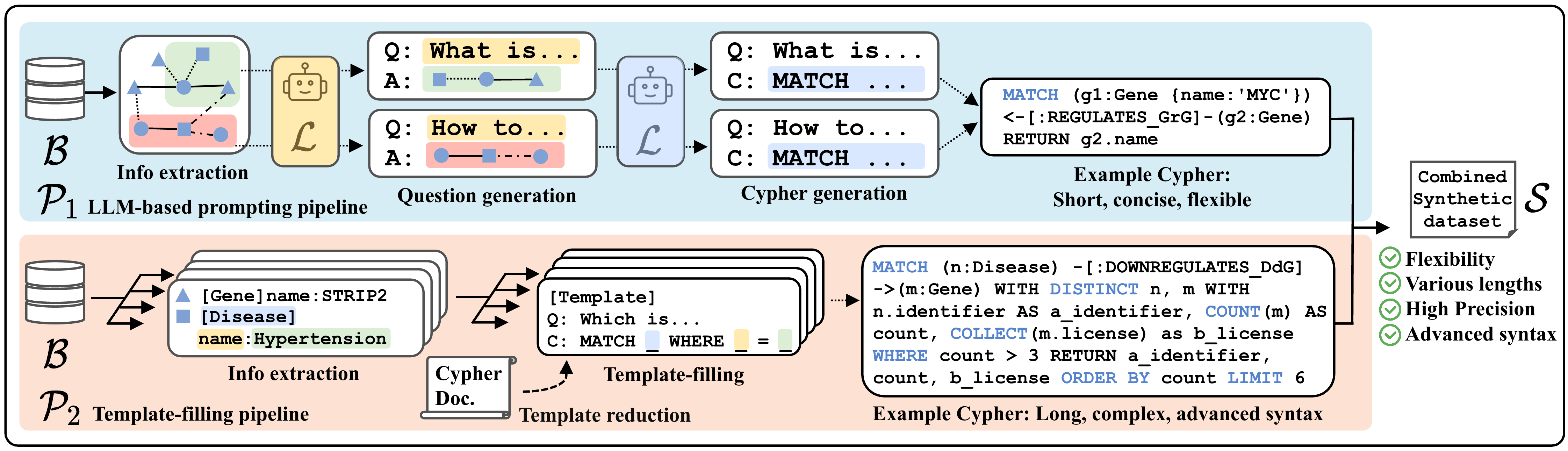}
  \caption {Workflow of two pipelines inside SyntheT2C.}
  \label{fig:main}
\end{figure*}

\subsection{Preliminaries}
The goal of the Text2Cypher task is to automatically translate a query $q$ written in natural language to corresponding Cypher query $c$. With the proposed pipelines $\mathcal{P}_1$ and $\mathcal{P}_2$, a synthetic dataset $\mathcal{S}$ is built to fine-tune the backbone LLM $\mathcal{L}$. The synthetic data is generated and validated with a Neo4j database $\mathcal{B}$ and a series of automatic validators $\mathcal{V}=[\mathcal{V}_1, \mathcal{V}_2, ..., \mathcal{V}_5]$. The synthetic dataset after all the validations is denoted as $\mathcal{S}_v$. Using $\mathcal{S}_v$, $\mathcal{L}$ is fine-tuned into $\mathcal{L}_{ft}$ . The Cypher queries generated by $\mathcal{L}$ (resp. $\mathcal{L}_{ft}$) are noted as $c_1$ (resp. $c_2$).

\subsection{Synthetic dataset generation}
Generating the synthetic dataset is not trivial because synthetic data usually has difficulty in balancing grammatical correctness, semantic correctness, node coverage, edge coverage, and Cypher complexity. As a result, we propose a method of generation with two pipelines, as illustrated in Figure \ref{fig:main}). The LLM-based prompting pipeline ($\mathcal{P}_1$), emphasizes semantic variety, while the template-filling pipeline ($\mathcal{P}_2$), focuses on syntactic complexity. By employing these complementary pipelines, we aim to produce a synthetic dataset that captures the nuanced balance of linguistic, semantic, and structural properties.

\subsubsection{LLM-based prompting pipeline}
\label{sec:llm-based prompting pipeline}
This pipeline adopts an idea similar to Knowledge Distillation: we use the Cyphers generated by a stronger LLM to SFT weaker LLMs. Half of $\mathcal{S}$ is built by few-shot prompting GPT-4o \citep{gpt4}. To simplify the process and ensure a higher quality of the generated data, we split the whole generation task into (1) extracting information from the database; (2) determining the question categories; and (3) generating the Cyphers for each category with extracted information.

The workflow for the LLM-based prompting method is delineated in Figure \ref{fig:main} (upper part, $\mathcal{P}_1$). Initially, we commence by extracting metadata from the KG stored in the Neo4j database $\mathcal{B}$. This extraction includes sampling example nodes and edges to construct few-shot prompts, along with capturing the schema of the database to facilitate the generation of grounded Cyphers. An illustrative instance of extracted metadata is provided in Appendix \ref{sec:example_kg_info}. Subsequently, this metadata serves as a foundational component in all ensuing prompts, ensuring the generation of executable Cyphers. Before initiating the Cypher generation process, a preliminary step involves prompting the LLM to propose potential question categories, thereby mitigating the risk of redundant outputs. The backbone LLM undergoes multiple iterations to propose these question categories, as detailed in the prompt showcased in Appendix \ref{sec:prompt_propose_categories}. These proposed categories are then consolidated to eliminate duplicates, as instructed in the prompt outlined in Appendix \ref{sec:prompt_merge_categories}. After the deduplication, GPT-4o is prompted to generate synthetic Question-Cypher pairs with the prompt outlined in Appendix \ref{sec:prompt_generate_questions}. In our experiment, we fix a list of 12 categories (referred to as \colorbox{myLightGray}{\texttt{categories}}) to facilitate the comparison.

\subsubsection{Template-filling pipeline}
The second pipeline of Cypher generation adopts the template-filling method, a classic approach in code generation known for its flexible output and potentially complex syntax. We introduce this pipeline as a complement to the first one, leveraging manually crafted templates to generate Cyphers with more advanced syntax, thereby enabling backbone $\mathcal{L}$ to solve complicated questions.

In this pipeline, depicted in Figure \ref{fig:main} (lower part, $\mathcal{P}_2$), numerous templates are initially manually authored. Subsequently, actual values from different fields are sampled from the Neo4j database $\mathcal{B}$ to populate these templates, resulting in the generation of complete executable Cypher queries.

One such template is illustrated in Figure \ref{fig:template_example}. In this example, the \colorbox{myLightGray}{\texttt{subschema}} is introduced to manage cases where the entire database cannot be loaded at once, necessitating the selection and injection of only the relevant subgraph into the prompt. The variables \colorbox{myLightGray}{\texttt{label\_i}} and \colorbox{myLightGray}{\texttt{prop\_j}} represent the randomly sampled names of nodes and their attributes. These templates are initially crafted taking inspiration from \textit{Cypher Generator} \citep{cypher_gen}, then enriched and verified by the authors. More details about the construction process of the templates are presented in Appendix \ref{sec:details_templates}.

\begin{figure*}[t]
  \includegraphics[width=\linewidth]{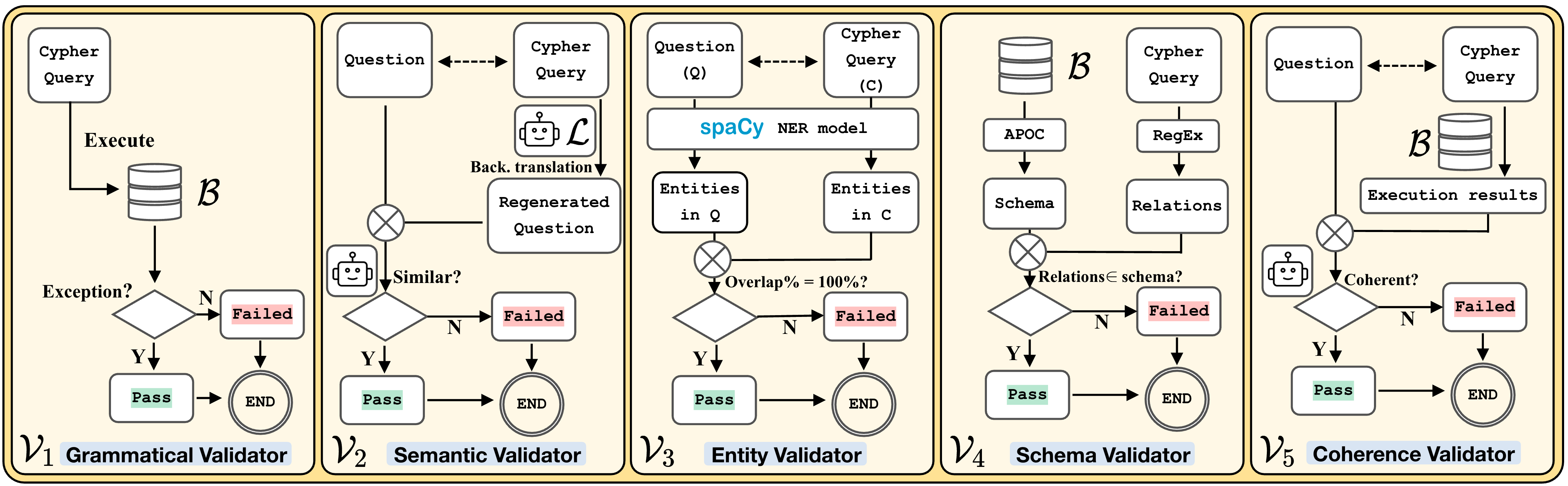}
  \caption {Illustration of the automatic validators.}
  \label{fig:validators}
\end{figure*}

\begin{figure}[ht]
\lstset{language=Python,
    basicstyle=\ttfamily,
    keywordstyle=\bfseries,
    showstringspaces=false,
    morekeywords={include, printf, prompt, question, schema, cypher, prompter}
}
\begin{lstlisting}[breaklines=true, breakatwhitespace=true, basicstyle=\scriptsize\ttfamily, columns=fullflexible]
def prompter(label_1, prop_1, prop_2):
    subschema = get_subgraph_schema(jschema, [label_1], 2, True)
    message = {
        "prompt": "Convert the following question into a Cypher query using the provided graph schema!",
        "question": f"""Find all {prop_1} for {label_1} that have {prop_2} after January 1, 2020!""",
        "schema": f"Graph schema: {subschema}",
        "cypher": f"MATCH (n:{label_1}) WHERE date(n.{prop_2}) > date('2020-01-01') RETURN n.{prop_1}"    }
    return message
\end{lstlisting}
\caption{Example template in Template-filling pipeline.}
\label{fig:template_example}
\end{figure}

Once these templates are established, synthetic Cyphers with complex syntax can be effortlessly generated. However, it is important to note that crafting and validating these templates require considerable time and effort.

\subsection{Quality validation}
To ensure the quality of the generated synthetic Question-Cypher pairs before their application in SFT, it's imperative to conduct thorough validation. However, manually scrutinizing thousands of Cypher queries is arduous and time-consuming. In response, a suite of automatic validators are proposed to alleviate the burden of manual inspection. In the end, the Cyphers that pass through these automated validators undergo a final round of meticulous manual validation by researchers.

\subsubsection{Automatic validation}
We propose five automatic validators: the Grammatical Validator, Semantic Validator, Entity Validator, Schema Validator, and Coherence Validator, each playing a crucial role in ensuring the integrity of the generated synthetic data. These validators' fundamental concepts are illustrated in Figure \ref{fig:validators}. The LLM used in the validators is GPT-3.5-Turbo.

The \textbf{Grammatical Validator} validates the syntax correctness of each Cypher in $\mathcal{S}$ by executing them in the deployed graph database $\mathcal{B}$. If a Cypher is executed without encountering any ``Error/Exceptions'', it is deemed to have passed this validation.

The design of \textbf{Semantic Validator} is inspired by the research in machine translation \citep{backtranslation}. This validator utilizes an LLM to translate the generated Cypher back into a natural language question. It then computes the semantic similarity between the translated question and the original question. If the similarity score exceeds a predefined threshold, the Cypher passes validation. We also tested an alternative implementation, where the LLM assesses semantic similarity directly. Both versions produce coherent validation results, with the latter being adopted for efficiency in subsequent experiments. The prompt used in this validator is presented in Appendix \ref{sec:prompt_semantic_validator}.

The \textbf{Entity Validator} assesses the coverage of entities in the generated Cyphers. The entities in the original question $q$ are extracted via Named Entity Recognition (NER) using the spaCy \citep{spacy} model \colorbox{myLightGray}{\texttt{en\_core\_web\_sm}}. Entities in the generated Cypher $c$ are parsed and extracted using Regular Expressions. A successful validation requires 100\% coverage of $q$'s entities in $c$. English entities are first transformed into lemmas using spaCy for fuzzy matching.

Subsequently, the \textbf{Schema Validator} ensures the correctness of relations in the generated Cyphers. Relations in $c$ are extracted via Regular Expressions and validated against the schema of $\mathcal{B}$. A Cypher passes this validation only when all contained relations are valid edges.

Lastly, the \textbf{Coherence Validator} executes the Cypher against $\mathcal{B}$ and evaluates the coherence between the execution results and the original question with LLM (with prompt in Appendix \ref{sec:prompt_coherence_validator}).

In the end, only Cyphers that have passed all validations proceed to manual validation.

\subsubsection{Manual validation}
Each Cypher checked by the validators is randomly assigned to two researchers, who independently assess its quality. If both researchers provide a unanimous judgment, their consensus is adopted. In cases of divergent opinions, a third researcher is brought in for further review. The final validation outcome for such Cyphers is determined through a majority vote among the three researchers. Over 98\% of the pairs passed the manual validation. Manual expertise involved is marginal as only less than 2\% of the pairs failed the manual validation.

\section{Experiments}
\subsection{LHY and Hetionet Graph databases}
In our experiment, we employed two Neo4j databases of general medical knowledge: the LHY Medical Knowledge Database (referred to as ``LHY'') and the Hetionet Medical Knowledge Database (referred to as ``Hetionet''). Both databases are publicly accessible, differing primarily in language: LHY is written in Chinese, whereas Hetionet is written in English. Their detailed statistics are presented in Appendix \ref{sec:statistics_lhy_hetionet}.

The LHY Database \citep{lhy} serves as the backend database for a Medical Question-Answering system. This database comprises comprehensive medical knowledge, encompassing a wide array of diseases, symptoms, drugs, and related information. Its content is sourced from medical websites, meticulously cleaned, reorganized, and stored within a Neo4j database. There are about 44k entities and 300k relations in it.

Hetionet \citep{rephetio} is an open and free-to-use database of biomedical knowledge resource implementing ``hetnet'' model. Aggregating insights from 29 public databases, Hetionet boasts a knowledge network spanning various fields, encompassing a wide array of entities, including genes, compounds, anatomical structures, diseases, symptoms, side effects, etc. There are approximately 47k entities and 2.2 million relations in the Hetionet database.

\begin{figure*}[t]
  \includegraphics[width=\linewidth]{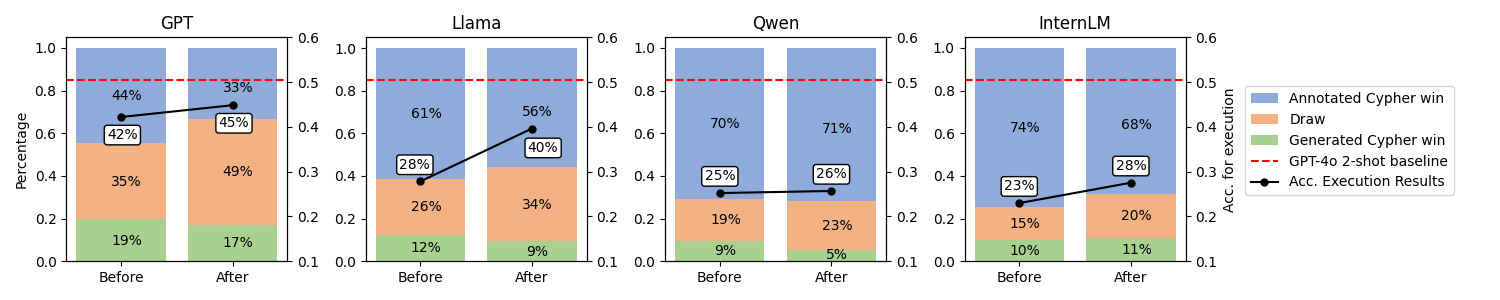}
  \caption {Result of Supervised Fine-Tuning each LLMs with MedT2C. Accuracy annotations marked in white box.}
  \label{fig:sft_result}
\end{figure*}
\label{sec:databases}

\subsection{Evaluation dataset and metrics}
\label{sec:eval_metrics}
We utilize a dataset comprising 300 manually annotated and verified samples to evaluate our experiments. This dataset includes 150 questions annotated based on the Hetionet and LHY databases, respectively. Take Hetionet as an example, for every category among the 12 categories generated in Section \ref{sec:llm-based prompting pipeline}, we employ GPT-3.5-Turbo to generate 10 new questions, forming 120 ``in-domain'' questions. Additionally, we introduce 3 unseen categories and generate 10 new questions for each new category, totaling 30 ``out-of-domain'' questions. For each of the 300 questions, the authors write a ground-truth Cypher query and test them manually to get the ground-truth execution results.

This annotated dataset allows us to evaluate two aspects of LLMs' Cypher generating performance: 

(1) \textbf{Cypher quality}, which is crucial if the generated Cypher is integrated into larger systems; 

(2) \textbf{Execution result accuracy}, to gauge the quality of the output for end users.

\subsubsection{Evaluation of Cypher quality}
The backbone LLMs, both pre-SFT and post-SFT, are tasked with generating Cyphers for the 300 questions in the evaluation dataset. Using GPT-4o \citep{gpt4}, we determine the superior Cypher from the two provided versions. For each pair of Cyphers, we conduct two evaluations by varying the order of presenting the Cypher queries in the prompt to mitigate order-induced bias. If evaluations of both orders yield identical results, this judgment is accepted as the final outcome; otherwise, it is deemed a ``Draw''.

\subsubsection{Evaluation of execution result accuracy}
The generated Cyphers $c_2$ are executed on database $\mathcal{B}$ to get execution results $res_{gen}$. Then the accuracy ($acc$) is calculated with the ground-truth execution results $res_{gt}$ like this:
\begin{equation}
    acc = \frac{\#(res_{gen} \cap res_{gt})}{\#(res_{gen})},
\end{equation}
where $\#(.)$ calculates the cardinality of a set.

\subsection{Experiment setup}
\subsubsection{Cypher LLMs (baselines)}
Extensive experiments are conducted with four LLMs, including open/closed-source models. For open-source models, we evaluate Llama3, Qwen2 and InternLM2. For closed-sources model, we test GPT. The exact versions of the backbone LLMs are listed in Appendix \ref{sec:llm_versions}. Our experiments focus on the performance boost brought by the SFT, therefore, the pre-SFT vanilla LLMs are used as the natural baselines. Empirically, the vanilla LLMs of the parameter size around 7B perform poorly on Cypher writing task. Only about 15\%-20\% of the few-shot generated Cyphers are executable, and the rate of semantically correct Cyphers is even lower. Thus, the few-shot prompting has become a default choice for 7B-level LLMs on Cypher writing task, and all the prompts we used in the experiments are 2-shot prompts. 

We selected these 7B level models for our experiments to highlight the effectiveness of our method to enhance the smaller models with the synthetic data constructed with larger models.

On top of this, we also report the performance of GPT-4 as a reference: when tested on the Evaluation dataset, GPT-4 achieves the averaged Execution Result Accuracy of 49.07\% (zero-shot) and 50.42\% (2-shot).

\subsubsection{Supervised Fine-Tuning} 
We utilize Low-Rank Adaptation (LoRA) to fine-tune the vanilla LLMs. Specifically, the open-source models are trained for 6 epochs with a linear scheduler, starting at a learning rate of 1e-6. AdamW is used as the optimizer, and the training batch size is 6. The fine-tuning of GPT is facilitated by its official API. The experiments on all LLMs, are conducted on Nvidia GeForce 4090 GPU. All the experiments totaled about 1100 GPU hours.

\subsection{Supervised Fine-Tuning experiments}
The backbone LLMs are fine-tuned with the MedT2C dataset. MedT2C contains 3000 samples in total, with 750 samples generated from each combination of the two pipelines and the two Neo4j databases. The MedT2C dataset contains high-quality Question-Cypher pairs that passed all the automatic validations as well as the manual validation. In Appendix \ref{sec:passing_rate} we report the passing rate of each validator as a guide for further improvement of MedT2C's data quality.

A list of LLMs including GPT, Llama, Qwen, and InternLM are fine-tuned using MedT2C. We evaluated the change in Cypher writing performance of these LLMs, and the results are shown in Figure \ref{fig:sft_result}. The results show that MedT2C helps the LLM to produce more Cypher queries that are on par with or better than the human annotated ones.

\begin{figure}[t]
  \includegraphics[width=\columnwidth]{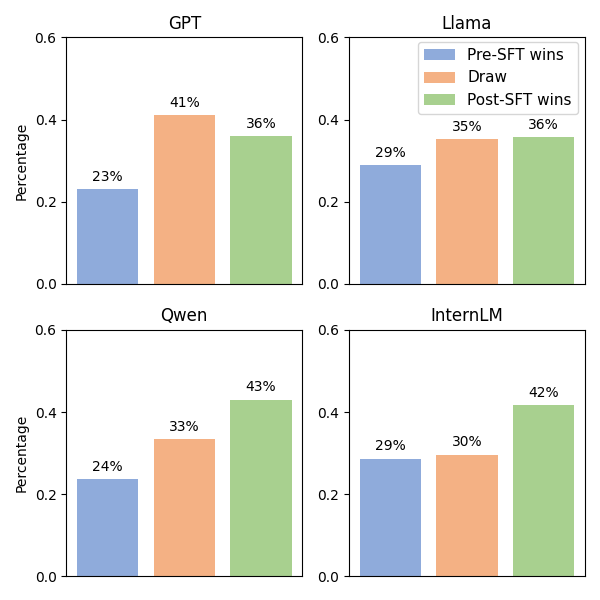}
  \caption{Impact of SFT on each LLM. The Cypher generated with pre-SFT and post-SFT LLMs are compared directly with GPT-4o.}
  \label{fig:sft_result_2}
\end{figure}

In Figure \ref{fig:sft_result}, the win rates are calculated in comparison with the ground-truth Cyphers. We further conduct an experiment to compare directly the $c_1$ and $c_2$ generated with the same LLM with GPT-4o, using the prompt presented in Appendix \ref{sec:prompt_cypher_quality_eval}. The comparison results are shown in Figure \ref{fig:sft_result_2}. From these results, we can conclude that while the improvement may appear minor when comparing against the ground-truth Cyphers, a visible enhancement in Cypher quality is evident when comparing to the Cyphers generated by the pre-SFT model. We explain this difference as follows: the human annotations have a far higher quality than the Cyphers generated by vanilla LLMs. Therefore, even though the LLMs are enhanced after SFT, their output is still inferior to the human-annotated Cyphers, which is why the evaluation results in Figure \ref{fig:sft_result} seem largely unchanged.

\subsection{Scaling experiments}
\begin{figure}[t]
  \includegraphics[width=\columnwidth]{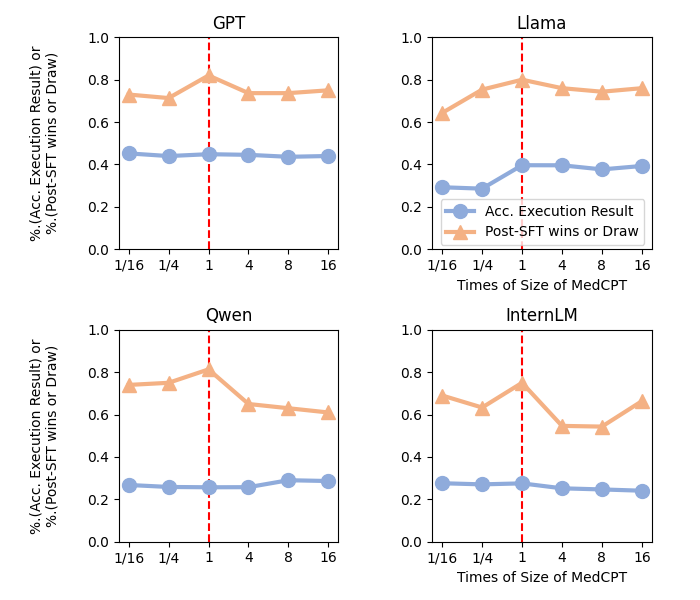}
  \caption{Plots of scaling test's results.}
  \label{fig:scaling_test}
\end{figure}

In this section, we test the scalability of our pipeline for generating synthetic data. We rerun the data generation pipelines to create scaled versions that are 1/16, 1/4, 4, 8, and 16 times the size of the original MedT2C. Vanilla LLMs are then fine-tuned with these scaled datasets. The results are reported in Figure \ref{fig:scaling_test}. These results demonstrate that, up to the size of the MedT2C dataset, increasing the size of the synthetic dataset leads to better performance, especially in terms of Cypher Quality. However, once the size exceeds that of MedT2C, further increasing the dataset size results in either marginal improvements or decreases. Based on this experiment, we determine the optimal size for the published MedT2C dataset (highlighted in red), as it balances efficiency and performance.

\subsection{Ablation experiments}
To evaluate the efficacy of each component introduced, we conduct a series of ablation experiments. First, we test the pipelines by running SFT experiments using only the data generated by each pipeline individually. We then verify the effectiveness of each automatic validator by evaluating them in isolation, using only one validator at a time. Since each component is designed to be modular and independent, we adopt this mode of ablation, rather than removing the components one by one from the complete setting, to emphasize the increment brought by each component separately. For both ablation tests, the backbone LLM is fixed as Llama3. The dataset size is set to be the same as MedT2C (3000 in total). The experiments results are reported in Table \ref{tab:ablation_res_1} and Table \ref{tab:ablation_res_2} respectively. Here the Cypher Quality is calculated with respect to ground-truth Cyphers. 

\begin{table}
  \centering
  \small
  \begin{tabular}{lcc}
    \hline
    \textbf{Settings} & \textbf{Cypher Quality} & \textbf{Result Acc.} \\
    \hline
    Pre-SFT   &38.67\%(--)        &27.83\%(--)   \\
    \quad LHY-LLM   & 41.67\%(+3.00)&27.86\%(+0.03)\\
    \quad LHY-Temp. & 34.67\%(-4.00)   &26.54\%(-1.29)\\
    \quad Hetionet-LLM   & 42.83\%(+4.16)&33.09\%(+5.26)\\
    \quad Hetionet-Temp. & 36.00\%(-2.67)&26.68\%(-1.15)\\
    \quad only LLM & 41.00\%(+2.33) & 29.02\%(+1.19)\\
    \textbf{All (MedT2C)} & \textbf{44.00\%}(+5.33)  & \textbf{39.65\%}(+11.82) \\
    \hline
  \end{tabular}
  \caption{Results of pipeline ablation test.}
  \label{tab:ablation_res_1}
\end{table}

As presented in Table \ref{tab:ablation_res_1}, when we use only the data generated by the template-filling pipeline to SFT the Llama3 model, the post-SFT performance actually declines. The reasons are: (1) template-filling pipeline generates rigid Cyphers which are easier to overfit; (2) when SFT with only template-filling data, LLMs tend to produce unnecessarily complicated Cypher queries (e.g., breaking one query into two and then merging them). From a global perspective, using data from both pipelines (marked as ``All'') can enhance the LLM's generalization capacity, as shown in comparison with the result obtained by fine-tuning only using the data generated with pipeline 1 (marked as "only LLM"). In other words, even though the template-filling data seems to have a negative impact, adding them to the fine-tuning dataset can further improve the overall Cypher writing performance.

\begin{table}
  \centering
  \small
  \begin{tabular}{lcc}
    \hline
    \textbf{Settings} & \textbf{Cypher Quality} & \textbf{Result Acc.} \\
    \hline
    Pre-SFT           &38.67\%(--)          &27.83\%(--)    \\
    \quad No validator      & 38.34\%(-0.33)  &27.96\%(+0.13)    \\
    \quad \checkmark Grammar V.   & 38.34\%(-0.33)  &28.95\%(+1.12)    \\
    \quad \checkmark Semantic V.  & 43.67\%(+5.00)  &31.65\%(+3.82)    \\
    \quad \checkmark Entity V.    & 40.00\%(+1.33)  &28.03\%(+0.20)    \\
    \quad \checkmark Schema V.    & 42.00\%(+3.33)  &26.11\%(-1.72)    \\
    \quad \checkmark Coherence V. & 41.33\%(+2.66)  &32.05\%(+4.22)    \\
    \textbf{All (MedT2C)} & \textbf{44.00\%}(+5.33)  & \textbf{39.65\%}(+11.82) \\
    \hline
  \end{tabular}
  \caption{Results of validator ablation test.}
  \label{tab:ablation_res_2}
\end{table}

Table \ref{tab:ablation_res_2} shows that each individual validator contributes some improvement, either in terms of Cypher quality or the accuracy of the execution results. Notably, the combination of all five validators yields the most significant increase in performance. This can be attributed to the validators' collective ability to mitigate the majority of the bugs in the SFT dataset, thereby enhancing the overall quality of the generated Cypher queries.

\section{Limitations}
The primary limitation of our work is the challenge in writing the templates. Besides, some adaptation work is necessary when applying them to new Neo4j databases (more details about adaptation to new database are discussed in Appendix \ref{sec:adaptation_templates}). Also, a portion of the generated Cypher queries are directly filtered out during the construction of MedT2C, which is not the most efficient solution. Finally, our current work focuses only on Cypher query language, similar method can be applied on other structured query languages in future research.

\section{Potential risks}
Though SyntheT2C is designed to automatically generate synthetic datasets, its usage requires close monitoring to prevent the inadvertent inclusion of private or sensitive information. Additionally, there is a slight residual risk for post-SFT LLMs of producing endless embedded Cyphers, which could potentially lead to issues such as Out-of-Memory.

\section{Conclusion}
We present SyntheT2C, a comprehensive framework to generate synthetic data for SFT various LLMs on the Text2Cypher task. Our approach encompasses dataset construction, data validation, and SFT evaluation, providing a reference framework for future research in the Cypher-related field. Additionally, our findings confirm the effectiveness of synthetic data, suggesting that similar techniques can address problems where annotation is difficult or insufficient. Finally, we will also open-source the MedT2C dataset, aiming to contribute to the technical advancements in relevant topics. 

\section*{Acknowledgements}
We would like to express our gratitude to all the supervisors and colleagues at Shanghai Artificial Intelligence Laboratory for their invaluable insights, feedback, and support throughout the research process. We also thank the reviewers for their constructive comments, which greatly improved the quality of this paper.

\clearpage
\bibliography{custom}

\clearpage
\appendix

\section{Example of extracted KG information}
\label{sec:example_kg_info}
Here we present the information (metadata) extracted from the KG database ``Hetionet'' in Figure \ref{fig:metadata_hetionet}. We stored the metadata of the KG, including the node properties, the relationship properties, and the valid relationships. This information is integrated in the following prompts to ensure that the LLM output is correct Cyphers. In other prompts, this metadata is referred to as \colorbox{myLightGray}{\texttt{schema}}.

\section{Prompts for LLM-based prompting pipeline}
\label{sec:llm_pipeline_prompts}
In this appendix, we present all the prompts we used in the LLM-based prompting pipeline.

\subsection{Prompt to propose categories of questions} 
\label{sec:prompt_propose_categories}
In Figure \ref{fig:prompt_generate_categories} we show the prompt used to propose candidate categories of questions. We decided to first generate categories of questions instead of generating directly the questions because this practice helps reduce duplicated questions.

\subsection{Prompt to generate questions for each category}
\label{sec:prompt_generate_questions}
This prompt presented in Figure \ref{fig:prompt_generate_questions} is used to generate questions in natural language for each proposed \colorbox{myLightGray}{\texttt{category}}. This prompt includes few-shot examples to help ensure the output Cypher follows the format requirements.

\subsection{Prompt to merge categories of questions}
\label{sec:prompt_merge_categories}
The prompt presented in Figure \ref{fig:prompt_merge_categories} is used to merge the previously generated categories. The merged and de-duplicated list of categories is then stored and will be referred to as \colorbox{myLightGray}{\texttt{category}} in later prompts.

\section{Details about constructing the templates}
\label{sec:details_templates}
Our templates are based on the 60 templates introduced in \textit{Cypher Generator} \citep{cypher_gen}. These templates were originally written for an older version of Neo4j, so the authors first correct and update them to match the syntax of Neo4j 5.13. Additionally, the authors refer to the official Neo4j documentation to create 20 new templates that cover the new syntax and new features in Neo4j 5.13. All the 80 templates are tested by filling them with sampled values and executing them manually. After testing, all templates are confirmed to be able to generate executable Cypher queries.

\section{Prompts used in automatic validators}
\subsection{Prompt of Semantic Validator}
\label{sec:prompt_semantic_validator}
Here we present the prompt used in the Semantic Validator in Figure \ref{fig:prompt_semantic_validator}. The \colorbox{myLightGray}{\texttt{schema}} mentioned in this prompt is the metadata presented in Appendix \ref{fig:metadata_hetionet}. The \colorbox{myLightGray}{\texttt{example}} represents the few-shot examples written by the authors, here we show the English example for the Hetionet database in Figure \ref{fig:english_example}. Lastly, the \colorbox{myLightGray}{\texttt{json\_object}} in the prompt contains the question and the Cypher query to be evaluated.

\subsection{Prompt of Coherence Validator}
\label{sec:prompt_coherence_validator}
In this appendix, we present the prompt used in the Coherence Validator in Figure \ref{fig:prompt_coherence_validator}. Similar to other prompts, we provided few-shot examples in this prompt. The \colorbox{myLightGray}{\texttt{question}} and \colorbox{myLightGray}{\texttt{results}} in the prompt are the original question and execution results used as the input for this validation.

\section{Important statistics of the LHY and the Hetionet databases}
\label{sec:statistics_lhy_hetionet}
Here we present the important statistics of the LHY database in Table \ref{tab:lhy_entity} and Table \ref{tab:lhy_relation}, including the examples of nodes and entities inside this database. The examples in both tables are translated from Chinese to English. Similarly, the important statistics of the Hetionet database with examples of nodes and entities are grouped in Table \ref{tab:hetionet_entity} and Table \ref{tab:hetionet_relation}.

\section{Exact versions of the backbone LLMs}
\label{sec:llm_versions}
The exact versions of the LLMs used in our experiments are listed in Table \ref{tab:llm_version_table}. Except GPT-3.5-Turbo, the backbone LLMs are deployed locally using the versions available on HuggingFace.

\section{Passing rate of MedT2C for each automatic validator}
\label{sec:passing_rate}
The passing rate of MedT2C dataset for each automatic validator is reported in Table \ref{tab:passing_rate}. The LLM used in the Semantic Validator and the Coherence Validator is GPT-3.5-Turbo. These two validators are not run on the LLM-based prompting pipeline because this pipeline uses GPT-4o. Given that GPT-4o is more powerful than GPT-3.5-Turbo, it is not accurate to evaluate its output with GPT-3.5-Turbo, nor with GPT-4o itself. Besides, noted that the passing rate of Coherence Validator is especially low compared to other passing rate. This is because for Coherence Validator specifically, the samples that failed any one of the previous validators is judged as \colorbox{myLightGray}{\texttt{False}} directly to save the calling of GPT API. Therefore the passing rate of Coherence Validator reported here is lower than the actual one, but it does not affect the ``All passed'' ratio.

\section{Prompts used for Cypher quality evaluation}
\label{sec:prompt_cypher_quality_eval}
We use GPT-4o to judge the quality of two versions of Cypher queries corresponding to the same set of questions written in natural language. The prompt used for this part is shown in Figure \ref{fig:prompt_cypher_quality}. We provide different aspects of evaluation and ask GPT-4o to give detailed reasons when evaluating because these techniques bring more accurate evaluation results in practice.

\section{Adaptation of templates to new database}
\label{sec:adaptation_templates}
The challenge of adapting a template-based pipeline to unseen databases is a known issue as described in the section of Limitations. It is even a common drawback of all template-based methods. However, we have made efforts to mitigate this inconvenience. Our templates are written based on the work Cypher Generator \citep{cypher_gen} and enriched referring to the official documentation of Cypher Query Language, aiming to cover all syntax patterns of Cypher. When adapting to a new database, we comment out the templates involving non-existent data types (e.g., there is no DATE data in LHY database, so we comment out the DATE-related templates when applying on LHY database). Similarly, developers can comment or un-comment these templates when applying them to other databases, using these templates directly without rewriting from scratch. This method has proven to be effective when we adapted the pipeline to Hetionet and LHY databases during our experiments.


\begin{figure*}[]
\begin{lstlisting}[breaklines=true, breakatwhitespace=true, basicstyle=\small\ttfamily, columns=fullflexible]
Node properties are the following:
Disease {easy_get: STRING, cure_lasttime: STRING, cured_prob: STRING, name: STRING, desc: STRING, prevent: STRING, cure_way: LIST, cause: STRING, cure_department: LIST},Drug {name: STRING},Food {name: STRING},Check {name: STRING},Department {name: STRING},Producer {name: STRING},Symptom {name: STRING}

Relationship properties are the following:
recommand_eat {name: STRING}, no_eat {name: STRING}, do_eat {name: STRING}, belongs_to {name: STRING}, common_drug {name: STRING}, drugs_of {name: STRING}, recommand_drug {name: STRING}, need_check {name: STRING}, has_symptom {name: STRING}, acompany_with {name: STRING}

The relationships are the following:
(:Disease)-[:belongs_to]->(:Department), (:Disease)-[:common_drug]->(:Drug), (:Disease)-[:recommand_drug]->(:Drug), (:Disease)-[:need_check]->(:Check), (:Disease)-[:has_symptom]->(:Symptom), (:Disease)-[:acompany_with]->(:Disease), (:Disease)-[:recommand_eat]->(:Food), (:Disease)-[:no_eat]->(:Food), (:Disease)-[:do_eat]->(:Food), (:Department)-[:belongs_to]->(:Department), (:Producer)-[:drugs_of]->(:Drug)
\end{lstlisting}
\caption{The metadata extracted from the Hetionet database.}
\label{fig:metadata_hetionet}
\end{figure*}

\begin{figure*}[]
\begin{lstlisting}[breaklines=true, breakatwhitespace=true, basicstyle=\small\ttfamily, columns=fullflexible]
You are an experienced and useful Python and Neo4j/Cypher developer.

I have a knowledge graph for which I would like to generate interesting questions that span 12 categories (or types) about the graph. They should cover single-node questions, two or three more nodes, relationships, and paths. Please suggest 12 categories together with their short descriptions. Here is the graph schema: 

{schema}
\end{lstlisting}
\caption{The prompt used to generate categories of questions.}
\label{fig:prompt_generate_categories}
\end{figure*}

\begin{figure*}[]
\begin{lstlisting}[breaklines=true, breakatwhitespace=true, basicstyle=\small\ttfamily, columns=fullflexible]
You are an experienced doctor and you have a knowledge graph for which you would like to generate interesting questions of 12 categories.

Here are some candidate categories: 

{categories_list}.

You should merge similar categories and remove the duplicates. Finally, give me a short description of each category.

\end{lstlisting}
\caption{The prompt used to merge proposed categories.}
\label{fig:prompt_merge_categories}
\end{figure*}

\begin{figure*}[]
\begin{lstlisting}[breaklines=true, breakatwhitespace=true, basicstyle=\small\ttfamily, columns=fullflexible]

You are an experienced Cypher developer and English-speaking doctor and a helpful assistant designed to output JSON

Generate {k} questions and their corresponding Cypher statements about the Neo4j graph database with the following schema:

{schema}

The questions should cover {category} and should be phrased in a natural conversational manner. Make the questions diverse and interesting. Make sure to use the latest Cypher version and that all the queries are working Cypher queries for the provided graph. You may add values for the node attributes as needed. Do not add any comments, do not label or number the questions. 

Here are some examples of the Question-Cypher pairs to be generated:

"question": "What are the diseases that commonly accompany 'Depression'?",
"cypher": "MATCH (d1:Disease {{name:'Depression'}}) -[:acompany_with]-> (d2:Disease) RETURN d2.name"

"question": "Can you list diseases that commonly accompany 'Cancer'?",
"cypher": "MATCH (d1:Disease {{name:'Cancer'}}) -[:acompany_with]-> (d2:Disease) RETURN d2.name",

Now it's your turn to generate the question and Cypher pairs:

\end{lstlisting}
\caption{The prompt used to generate questions.}
\label{fig:prompt_generate_questions}
\end{figure*}

\begin{table*}
  \centering
  \small
  \begin{tabular}{p{0.16\linewidth}lp{0.55\linewidth}}
    \hline
    \textbf{Ent. Type}  & \textbf{\# Ent.} & \textbf{Examples}\\
    \hline
    Check & 3,353  & \small{Bronchography}     \\
    Department & 54  & \small{Department of Plastic and Reconstructive Surgery}     \\
    Disease & 8,807 & \small{Thrombosed Vasculitis}       \\
    Drug & 3,828 & \small{Jingwanhong Hemorrhoid Cream}  \\
    Food & 4,870  & \small{Tomato and Vegetable Beef Ball Soup}    \\
    Producer & 17,201  & \small{Tongyi Pharmaceutical Penicillin V Potassium Tablets}    \\
    Symptom & 5,998 & \small{Hypertrophy of breast tissue}   \\
    \hline
    \textbf{Total} & \textbf{44,111}  & /  \\
    \hline
  \end{tabular}
  \caption{Entities in LHY Database.}
  \label{tab:lhy_entity}
\end{table*}

\begin{table*}
  \centering
  \small
  \begin{tabular}{p{0.16\linewidth}lp{0.55\linewidth}}
    \hline
    \textbf{Rel. Type}  & \textbf{\# Rel.} & \textbf{Examples}\\
    \hline
    belongs\_to & 8,844  & <Gynaecology, belongs\_to, Obstetrics and Gynaecology>     \\
    common \_drug & 14,649  & <Yang Qiang, common\_drug, Phentolamine mesylate dispersible tablets>  \\
    do\_eat & 22,238 & <Thoracic spine fracture, do\_eat, Blackfish> \\
    drugs\_of & 17,315 & <Penicillin V Potassium Tablets, drugs\_of, Tongyi Pharmaceuticals Penicillin V potassium tablets> \\
    need \_check & 39,422 & < Unilateral emphysema, need\_check, Bronchography>    \\
    no\_eat & 22,247  & <Lip disease, no\_eat, Almonds>  \\
    recommend\_drug & 59,467 & <Mixed hemorrhoids, recommend\_drug, Jingwanhong Hemorrhoid Cream>   \\
    recommend\_eat & 40,221 & <Synovial effusion, recommend\_eat, Beef Ball Soup with Tomato and Vegetable Punch> \\
    has\_ symptom & 5,998 & <Early Breast Cancer, has\_symptom, Hypertrophy of breast tissue>  \\
    accompany \_with & 12,029 & <Valvular insufficiency of the traffic veins of the lower extremities, accompany\_with, Thromboembolic vasculitis>   \\
    \hline
    \textbf{Total} & \textbf{294,149}  & /  \\
    \hline
  \end{tabular}
  \caption{Relations in LHY Database.}
  \label{tab:lhy_relation}
\end{table*}

\begin{table*}
  \centering
  \small
  \begin{tabular}{p{0.32\linewidth}lp{0.39\linewidth}}
    \hline
    \textbf{Ent. Type}  & \textbf{\# Ent.} & \textbf{Examples}\\
    \hline
    Anatomy & 402  & \small{Digestive System}     \\
    Biological\_process & 11,381  & \small{Protein Sialylation}      \\
    Cellular\_component & 1,391  & \small{Meiotic Spindle}     \\
    Compound & 1,552  &  \small{Mannitol}    \\
    Disease & 137  &  \small{Hypertension}    \\
    Gene & 20,945  & \small{STRIP2}     \\
    Molecular\_function & 2,884  & \small{Vitamin Transporter Activity}     \\
    Pathway & 1,822  & \small{Glycolysis}     \\
    Pharmacologic\_class & 345  & \small{Decreased Blood Pressure}     \\
    Side\_effect & 5,734  & \small{Subileus}     \\
    Symptom & 438  & \small{Ageusia}     \\
    
    \hline
    \textbf{Total} & \textbf{47,031}  & /  \\
    \hline
  \end{tabular}
  \caption{Entities in Hetionet Database.}
  \label{tab:hetionet_entity}
\end{table*}
 
\begin{table*}
  \centering
  \small
  \begin{tabular}{lll}
    \hline
    \textbf{Rel. Type}  & \textbf{\# Rel.} & \textbf{Examples}\\
    \hline
    Anatomy–downregulates–Gene & 102,240  & <Bronchus, downregulates, GRIA2>     \\
    Anatomy–expresses–Gene & 526,407  & <Myocardium, expresses, EFHD1>     \\
    Anatomy–upregulates–Gene & 97,848  & <Adipose tissue, upregulates, PARM1>     \\
    Compound–binds–Gene & 11,571  & <Sildenafil, binds, CYP3A4>     \\
    Compound–causes–Side\_Effect & 138,944  & <Ciprofloxacin, causes, Visual Disturbance>     \\
    Compound–downregulates–Gene & 21,102  & <Tacrolimus, downregulates, UBE2C>     \\
    Compound–palliates–Disease & 390  & <Fluvoxamine, palliates, Panic Disorder>     \\
    Compound–resembles–Compound & 6,486  & <Clotrimazole, resembles, Bifonazole>     \\
    Compound–treats–Disease & 755  & <Reserpine, treats, Hypertension>     \\
    Compound–upregulates–Gene & 18,756  & <Estriol, upregulates, KLHL9>     \\
    Disease–associates–Gene & 12,623  & <Parkinson's Disease, associates, HTR7>     \\
    Disease–downregulates–Gene & 7,623  & <Schizophrenia, downregulates, MLST8>     \\
    Disease–localizes–Anatomy & 3,602  & <Migraine, localizes, Brain>     \\
    Disease–presents–Symptom & 3,357  & <Lung Cancer, presents, Constipation>     \\
    Disease–resembles–Disease & 543  & <Bone Cancer, resembles, Head and Neck Cancer>     \\
    Disease–upregulates–Gene & 7,731  & <Malaria, upregulates, JAK2>     \\
    Gene–covaries–Gene & 61,690  & <IMP3, covaries, OR8U8>     \\
    Gene–interacts–Gene & 147,164  & <TRIM27, interacts, MED21>     \\
    Gene–participates–Biological\_Process & 559,504  & <ABCA1, participates, Lipid Homeostasis>     \\
    Gene–participates–Cellular\_Component & 73,566  & <KLHL14, participates, Neuronal Cell Body>     \\
    Gene–participates–Molecular\_Function & 97,222  & <TOP2B, participates, ATPase Activity>     \\
    Gene–participates–Pathway & 84,372  &  <GGT5, participates, Metabolism>    \\
    Gene-regulates-Gene & 265,672  & <BCCIP, regulates, HLTF>     \\
    Pharmacologic\_Class–includes–Compound & 1,029  & <Allergens, includes, Benzocaine>     \\
    
    \hline
    \textbf{Total} & \textbf{2,250,197}  & /  \\
    \hline
  \end{tabular}
  \caption{Relations in Hetionet Database.}
  \label{tab:hetionet_relation}
\end{table*}

\begin{figure*}[]
\begin{lstlisting}[breaklines=true, breakatwhitespace=true, basicstyle=\small\ttfamily, columns=fullflexible]
You are an experienced Cypher developer, English Master, and a helpful assistant that helps me to verify whether the cypher is coherent with the question!
You will be given a JSON object containing a question and a cypher query. You should first take a look at the schema provided below. The schema is the graph database on which the cypher queries will be run. 

The schema:

{schema}

You must organize your answer step by step and in the end, you should make your judgment.

Here are three areas that you should pay attention to:
1. whether the output of cypher is coherent with the question, which means that the output of cypher must contain the information that the question asks.
2. If the question points out a piece of key information, you should check whether this key information is pointed out in the cypher. For example, if the question provides a piece of exact information such as the exact name of the disease, this information can not be inconsistent in the cypher. If there is no exact key information, you can skip this area.
3. whether this cypher answers the question provided in the JSON object. You should simulate the cypher step by step according to the schema provided. Then you should judge whether this cypher is in line with the question.

You should make your judgment according to these three areas. If there are no problems in these three areas in the cypher, you must answer with 'True'. Otherwise, you should answer with 'False'.

Here are some example JSON objects:

{example}

Now it's your turn to answer! Here is the JSON object you should evaluate:

{json_object}

Now evaluate carefully the JSON object and provide your answer step by step.
\end{lstlisting}
\caption{The prompt used in Semantic validator.}
\label{fig:prompt_semantic_validator}
\end{figure*}


\begin{figure*}[]
\begin{lstlisting}[breaklines=true, breakatwhitespace=true, basicstyle=\small\ttfamily, columns=fullflexible]
<|Example 1|>
{
    "question": "Which diseases belong to the 'Psychiatry' department?",
    "cypher": "MATCH (d:Disease)-[:belongs_to]->(dept:Department) WHERE dept.name = 'Neurology' RETURN d.name"
},
<|Answer 1|>
The cypher is not in line with the question because the question is to find the diseases in the 'Psychiatry' department but the department name in the cypher is 'Neurology' department.
Since the key information is inconsistent, I would mark this JSON object as False.

<|Example 2|>
{
    "question": "Which foods should be avoided for the disease 'Brain tumor'?",
    "cypher": "MATCH (d1:Disease {name:'Brain tumor'})-[:no_eat]->(d2:Food) RETURN d2.name"
}
<|Answer 2|>
Firstly, the output of cypher contains the key information 'the food' asked by the question.
Secondly, the key information 'Brain tumor' provided in the question is contained in the cypher.
Finally, the logic of cypher is exactly similar to the question.
So, I think this JSON object is True.

<|Example 3|>
{
    "question": "What pathways do the genes 'BRCA1' and 'BRCA2' participate in?",
    "cypher": "MATCH (g:Gene)-[:PARTICIPATES_GpPW]->(:Pathway) WHERE g.name IN ['BRCA1', 'BRCA2'] RETURN g.name"
}
<|Answer 3|>
There are two errors.
Firstly, as the question asks for pathways but the output of cypher is the name of the gene, the output of the cypher is inconsistent with the question.
Secondly, the question is to find the pathway that both the genes 'BRCA1' and 'BRCA2' participate in. But the cypher matches the pathways that 'BRCA1' or 'BRCA2' participates in. The logic 'AND' and 'OR' are totally different.
Therefore, I think this JSON object is False.

\end{lstlisting}
\caption{The English few-shot examples used in the Semantic Validator.}
\label{fig:english_example}
\end{figure*}

\begin{figure*}[]
\begin{lstlisting}[breaklines=true, breakatwhitespace=true, basicstyle=\small\ttfamily, columns=fullflexible]
You are an experienced medical assistant who has mastered English and medical knowledge.
            
You will be given a question and the responses given by the doctor. The doctor is very professional, he gives direct responses. But he sometimes misunderstands the problem. Your task is to check if the results are coherent with the question by analyzing the category. 
For example, if the question asks for food and the answer is food, in this case, it is relevant because the category is the same. Even if the foods don't seem to be directly related, you can not deny them because the doctor is professional.
But if the question asks for food, the doctor gives the response on sports. You should point out this error because the category is different.

As a medical assistant, you just need to pay attention to whether the category of the answer corresponds to the category that the question asks. You don't need to think about the reasonableness of the answer.
Answer with 'True' if the category is the same. Otherwise, answer with 'False'.
You need to carefully explain your answer.

Here are some examples of questions and results:

<Example 1>
Question: Find out the diseases associated with the 'Oncology' department.
Responses by the doctor: Breast cancer, Pancreatic cancer, Colon cancer
Your reply: Breast cancer, pancreatic cancer, and colon cancer belong to the Oncology department. And the question asks for diseases. So I think it is relevant, and my answer is True.

<Example 2>
Question: Which foods should be avoided for the disease 'Coeliac disease'?
Responses by the doctor: Swimming, Running, Biking, Walking
Your reply: The responses are sports. But this question asks for food. So I think it is not relevant, my answer is False. 

Now it's your turn to verify if the responses are relevant to the question.
Remember! You just need to pay attention to whether the answer corresponds to the question. You don't need to think about the reasonableness of the answer.

Question:{question}
Responses by the doctor: {results}
Your reply:

\end{lstlisting}
\caption{The prompt used in Coherence Validator.}
\label{fig:prompt_coherence_validator}
\end{figure*}

\begin{table*}
    \small
    \centering
  \begin{tabular}{ccl}
    \hline
    \textbf{LLM name}  & \textbf{LLM version} & \textbf{LLM site} \\
    \hline
    GPT & gpt-3.5-turbo-16k  & https://platform.openai.com/docs/models/gpt-3.5-turbo   \\
    Llama3 & Meta-Llama-3-8B  & https://huggingface.co/meta-llama/Meta-Llama-3-8B-Instruct  \\
    InternLM2 & internlm2-7B & https://huggingface.co/internlm/internlm2-base-7b \\
    Qwen2 & Qwen2-7B   & https://huggingface.co/Qwen/Qwen2-7B     \\
    \hline
  \end{tabular}
  \caption{Versions of the backbone LLMs}
  \label{tab:llm_version_table}
\end{table*}

\begin{table*}
  \centering
  \small
  \begin{tabular}{>{\centering\arraybackslash}m{1.2cm}>{\centering\arraybackslash}m{3cm}>{\centering\arraybackslash}m{1.8cm}>{\centering\arraybackslash}m{1.4cm}>{\centering\arraybackslash}m{1.4cm}>{\centering\arraybackslash}m{1.4cm}>{\centering\arraybackslash}m{1.4cm}>{\centering\arraybackslash}m{1.4cm}}
  \hline
    \textbf{Database} & \textbf{Pipeline}            & \textbf{Grammatical Validator} & \textbf{Semantic Validator} & \textbf{Entity Validator} & \textbf{Schema Validator} & \textbf{Coherence Validator} & \textbf{All passed} \\
    \hline
    LHY      & LLM-based prompting & 99.69\%               & N/A                & 99.62\%          & 82.77\%          & N/A              & 83.87\%   \\
    LHY      & Template-filling    & 99.87\%               & 92.34\%            & 100\%            & 99.87\%          & 28.59\%          & 27.21\%   \\
    Hetionet & LLM-based prompting & 96.08\%               & N/A                & 99.08\%          & 61.69\%          & N/A              & 64.79\%   \\
    Hetionet & Template-filling    & 100\%                 & 91.81\%            & 99.52\%          & 100\%            & 38.15\%          & 36.66\%   \\
    \hline
    \end{tabular}
    \caption{MedT2C's passing rates of each automatic validator.}
  \label{tab:passing_rate}
\end{table*}

\begin{figure*}[]
\begin{lstlisting}[breaklines=true, breakatwhitespace=true, basicstyle=\small\ttfamily, columns=fullflexible]
You are an expert in medical field and Cypher query language. You are asked to evaluate the quality of the Cypher queries generated by 2 models for the same question. You will be first given the question written in natural language. Then you will be given the Cypher queries generated by 2 models. Your task is to compare the quality of these two Cyphers and select the better one. You should consider the following aspects when selecting the better Cypher: 
    1. Syntactical correctness: whether the Cypher query is syntactically correct; 
    2. Semantic correctness: whether the Cypher query can correctly answer the question; 
    3. Readability: whether the Cypher query is easy to read and understand; 
    4. Efficiency: whether the Cypher query is efficient in terms of time and space complexity; 
    5. Conciseness: whether the Cypher query is concise and clear; 
    6. Completeness: whether the Cypher query can cover all the necessary information in the database. 
    You should select the better Cypher query based on these aspects. Output your selected Cypher as well as your reasons.
        
Here is the question:
{
    "question": "{{ question }}"
}

Here are the outputs of the models:
[
    {
        "number": "1",
        "cypher": "{{ cypher_1 }}"
    },
    {
        "number": "2",
        "cypher": "{{ cypher_2 }}"
    }
]
Your output should be in the following format, DO NOT output anything other than this JSON object: 
{
"better_cypher": "1",
"reason": "reasons why 1 is selected"  
}

Now select the better Cypher and give your reasons:

\end{lstlisting}
\caption{The prompt used in Cypher quality evaluation.}
\label{fig:prompt_cypher_quality}
\end{figure*}

\end{document}